\documentclass[conference]{IEEEtran}
\IEEEoverridecommandlockouts

\usepackage{cite}
\usepackage{amsmath,amssymb,amsfonts}
\usepackage{algorithm}
\usepackage{algpseudocode}
\usepackage{graphicx}
\usepackage{textcomp}
\usepackage{xcolor}
\usepackage{booktabs}
\usepackage{multirow}
\usepackage{hyperref}
\usepackage{tikz}
\usetikzlibrary{shapes.geometric, arrows, positioning, calc, patterns}
\usepackage{pgfplots}
\pgfplotsset{compat=1.18}

\def\BibTeX{{\rm B\kern-.05em{\sc i\kern-.025em b}\kern-.08em
    T\kern-.1667em\lower.7ex\hbox{E}\kern-.125emX}}

\begin{document}

\title{Refining Graphical Neural Network Predictions Using Flow Matching for Optimal Power Flow with Constraint-Satisfaction Guarantee}

\author{
Kshitiz Khanal\\
Institute for Transportation Research and Education\\
North Carolina State University\\
Raleigh, NC 27695, USA\\
\texttt{kkhanal2@ncsu.edu}
}

\maketitle

\begin{abstract}
The DC Optimal Power Flow (DC-OPF) problem is fundamental to power system operations, requiring rapid solutions for real-time grid management. While traditional optimization solvers provide optimal solutions, their computational cost becomes prohibitive for large-scale systems requiring frequent recalculations. Machine learning approaches offer promise for acceleration but often struggle with constraint satisfaction and cost optimality. We present a novel two-stage learning framework that combines physics-informed Graph Neural Networks (GNNs) with Continuous Flow Matching (CFM) for solving DC-OPF problems. Our approach embeds fundamental physical principles—including economic dispatch optimality conditions, Kirchhoff's laws, and Karush-Kuhn-Tucker (KKT) complementarity conditions—directly into the training objectives. The first stage trains a GNN to produce feasible initial solutions by learning from physics-informed losses that encode power system constraints. The second stage employs CFM, a simulation-free continuous normalizing flow technique, to refine these solutions toward optimality through learned vector field regression. Evaluated on the IEEE 30-bus system across five load scenarios ranging from 70\% to 130\% nominal load, our method achieves near-optimal solutions with cost gaps below 0.1\% for nominal loads and below 3\% for extreme conditions, while maintaining 100\% feasibility. Our framework bridges the gap between fast but approximate neural network predictions and optimal but slow numerical solvers, offering a practical solution for modern power systems with high renewable penetration requiring frequent dispatch updates.
\end{abstract}


\section{Introduction}

\subsection{Motivation and Context}

The Optimal Power Flow (OPF) problem represents a cornerstone of modern power system operations, determining the most economical generator dispatch to meet electricity demand while satisfying operational constraints \cite{carpentier1962contribution}. As power grids integrate increasing levels of variable renewable energy sources, the need for frequent OPF solutions has intensified—grid operators may need to solve OPF problems hundreds or thousands of times per day to accommodate rapidly changing generation and load patterns \cite{li2024optimal}.

Traditional numerical optimization methods, while providing provably optimal solutions, face computational challenges in large-scale systems. Interior point methods and sequential quadratic programming can require seconds to minutes per solve \cite{nellikkath2022physics}, which becomes prohibitive when real-time decisions are needed. The DC Optimal Power Flow (DC-OPF) formulation, which linearizes power flow equations, offers a more tractable alternative but still presents computational bottlenecks for large systems with frequent updates \cite{gao2023physics}.

\subsection{Machine Learning for OPF: Promise and Challenges}

Recent years have witnessed growing interest in applying machine learning to accelerate OPF solutions \cite{zamzam2020learning, pan2020deepopf}. The core idea is to shift computational burden from online optimization to offline training: a neural network learns to map system states (loads, generation) to optimal dispatch decisions. Once trained, inference requires only a forward pass through the network, potentially offering significant speedup.

However, applying machine learning to OPF faces three critical challenges:

\textbf{1. Constraint Satisfaction:} Neural networks inherently produce unconstrained outputs. OPF solutions must satisfy numerous physical and operational constraints—power balance, generator limits, line flow limits—which standard neural networks frequently violate \cite{zeng2024qcqp}.

\textbf{2. Optimality Gap:} Even when feasible, neural network predictions often exhibit significant cost gaps compared to optimal solutions. Generic supervised learning objectives (e.g., mean squared error) do not directly optimize for economic dispatch quality \cite{chen2025physics}.

\textbf{3. Generalization:} OPF solutions must generalize across diverse operating conditions, including stress scenarios outside the training distribution. Standard data-driven approaches often fail when encountering unusual load patterns or generation contingencies \cite{li2025physics}.

\subsection{Physics-Informed Learning for Power Systems}

Physics-Informed Neural Networks (PINNs) offer a promising direction by embedding domain knowledge directly into learning objectives \cite{nellikkath2022physics, li2025physics}. Rather than purely mimicking input-output mappings, PINNs incorporate physical laws—such as Kirchhoff's current law, power flow equations, and optimality conditions—as regularization terms or hard constraints in the loss function. This approach has shown improved generalization and constraint satisfaction in various power system applications \cite{gao2023physics, wang2024data}.

For OPF specifically, the Karush-Kuhn-Tucker (KKT) optimality conditions provide a rigorous mathematical characterization of optimal solutions. At an optimal dispatch, marginal generation costs must equalize across generators (subject to active constraints), embodying the fundamental economic dispatch principle. Several works have explored incorporating KKT conditions into neural network training \cite{chen2025physics, liu2025physics}.

Graph Neural Networks (GNNs) present another natural fit for power system problems, as electrical grids possess inherent graph structure with buses as nodes and transmission lines as edges \cite{falconer2023leveraging, suri2025powergnn}. GNNs can leverage topological information through message passing, potentially improving generalization across different network configurations \cite{donon2020neural}.

\subsection{Continuous Flow Matching: A New Generative Modeling Paradigm}

Continuous Flow Matching (CFM) represents a recent breakthrough in generative modeling, offering an efficient alternative to diffusion models and traditional normalizing flows \cite{lipman2023flow, tong2023improving}. CFM trains continuous normalizing flows (CNFs) through a simulation-free objective, learning vector fields that transform samples from a simple distribution to a complex target distribution. Unlike diffusion models which require iterative denoising, CFM enables direct trajectory integration through learned ordinary differential equations (ODEs).

The key insight of CFM is regression of conditional vector fields along probability paths, circumventing expensive likelihood calculations and ODE integrations during training. This makes CFM particularly attractive for conditional generation tasks where we seek to refine initial predictions toward optimal solutions.

\subsection{Our Contributions}

We present a novel two-stage framework combining physics-informed GNNs with CFM for DC-OPF:

\begin{itemize}
\item \textbf{Physics-Informed GNN (Stage 1):} We develop a GNN architecture that encodes power system topology and trains with losses derived from fundamental physical principles: economic dispatch optimality (marginal cost equalization), power balance (Kirchhoff's law), generator limits, and KKT complementarity conditions. This produces feasible initial dispatches.

\item \textbf{Continuous Flow Matching Refinement (Stage 2):} We introduce CFM as a refinement mechanism, learning vector fields that transform initial GNN predictions toward optimal solutions. The CFM module trains on regression tasks mapping feasible solutions to optimal targets, then refines predictions through ODE integration at inference.

\item \textbf{Differentiable Projection Operators:} We design soft differentiable projection operators that maintain constraint feasibility during training while enabling smooth gradient flow, along with hard projections for inference.

\item \textbf{Comprehensive Evaluation:} We evaluate our framework across five distinct load scenarios (70\%-130\% nominal) on IEEE 30-bus system, demonstrating near-optimal performance ($<$0.1\% cost gap) for nominal conditions and robust performance ($<$3\% gap) even under extreme stress.
\end{itemize}

Our approach achieves a practical balance: the speed of neural network inference with constraint satisfaction and near-optimal economic dispatch quality approaching traditional solvers. The complete implementation and trained models are available online\footnote{\url{https://github.com/kshitizkhanal7/flow_refiner_dcopf}}.

\section{Problem Formulation}

\subsection{DC Optimal Power Flow}

The DC-OPF problem seeks the minimum-cost generator dispatch satisfying power balance and operational constraints. We formulate the problem as:

\begin{subequations}\label{eq:dcopf}
\begin{align}
\min_{\mathbf{p}_g} \quad & \sum_{i=1}^{N_g} C_i(p_{g,i}) \label{eq:objective}\\
\text{s.t.} \quad & \sum_{i=1}^{N_g} p_{g,i} = \sum_{j=1}^{N_b} p_{d,j} \label{eq:power_balance}\\
& p_{g,i}^{\min} \leq p_{g,i} \leq p_{g,i}^{\max}, \quad \forall i \label{eq:gen_limits}\\
& |\mathbf{P}_f| \leq \mathbf{P}_f^{\max} \label{eq:line_limits}
\end{align}
\end{subequations}

where:
\begin{itemize}
\item $\mathbf{p}_g \in \mathbb{R}^{N_g}$: generator active power outputs
\item $C_i(p_{g,i}) = c_{2,i} p_{g,i}^2 + c_{1,i} p_{g,i} + c_{0,i}$: quadratic cost function for generator $i$
\item $\mathbf{p}_d \in \mathbb{R}^{N_b}$: bus load demands (inputs)
\item $p_{g,i}^{\min}, p_{g,i}^{\max}$: generator capacity limits
\item $\mathbf{P}_f$: branch power flows (functions of $\mathbf{p}_g$ via DC power flow)
\item $N_g, N_b$: number of generators and buses
\end{itemize}

The DC approximation assumes: (1) voltage magnitudes near 1.0 p.u., (2) small angle differences, (3) negligible line resistance, enabling linear power flow equations:

\begin{equation}
\mathbf{P}_f = \mathbf{PTDF} \cdot (\mathbf{A}_g \mathbf{p}_g - \mathbf{p}_d)
\end{equation}

where $\mathbf{PTDF}$ is the Power Transfer Distribution Factor matrix and $\mathbf{A}_g$ is the generator-bus incidence matrix.

\subsection{KKT Optimality Conditions}

The Karush-Kuhn-Tucker conditions characterize optimal solutions to \eqref{eq:dcopf}. The Lagrangian is:

\begin{multline}
\mathcal{L} = \sum_{i=1}^{N_g} C_i(p_{g,i}) + \lambda \left(\sum_{j=1}^{N_b} p_{d,j} - \sum_{i=1}^{N_g} p_{g,i}\right) \\
+ \sum_{i=1}^{N_g} \mu_i^{\min}(p_{g,i}^{\min} - p_{g,i}) + \sum_{i=1}^{N_g} \mu_i^{\max}(p_{g,i} - p_{g,i}^{\max})
\end{multline}

At optimum $\mathbf{p}_g^*$, the following conditions hold:

\begin{subequations}\label{eq:kkt}
\begin{align}
\frac{\partial C_i}{\partial p_{g,i}}\bigg|_{\mathbf{p}_g^*} - \lambda + \mu_i^{\max} - \mu_i^{\min} &= 0, \quad \forall i \label{eq:kkt_stationarity}\\
\mu_i^{\min} \geq 0, \quad \mu_i^{\max} &\geq 0, \quad \forall i \label{eq:kkt_dual_feasibility}\\
\mu_i^{\min}(p_{g,i}^* - p_{g,i}^{\min}) &= 0, \quad \forall i \label{eq:kkt_complementarity_min}\\
\mu_i^{\max}(p_{g,i}^{\max} - p_{g,i}^*) &= 0, \quad \forall i \label{eq:kkt_complementarity_max}
\end{align}
\end{subequations}

\textbf{Economic Interpretation:} Equation \eqref{eq:kkt_stationarity} states that marginal costs must equal the system marginal price $\lambda$ (adjusted for binding constraints). For unconstrained generators:
\begin{equation}
\frac{\partial C_i}{\partial p_{g,i}} = 2c_{2,i}p_{g,i} + c_{1,i} = \lambda
\end{equation}

This equal incremental cost principle is fundamental to economic dispatch: all generators operating between limits should have equal marginal costs.

\subsection{Machine Learning Formulation}

We seek to learn a function $f_\theta: \mathbb{R}^{N_b} \rightarrow \mathbb{R}^{N_g}$ parameterized by neural networks that maps loads to near-optimal dispatch:
\begin{equation}
\hat{\mathbf{p}}_g = f_\theta(\mathbf{p}_d)
\end{equation}

The challenge is to design $f_\theta$ and its training procedure such that:
\begin{enumerate}
\item \textbf{Feasibility:} $\hat{\mathbf{p}}_g$ satisfies constraints \eqref{eq:power_balance}-\eqref{eq:gen_limits}
\item \textbf{Optimality:} $C(\hat{\mathbf{p}}_g) \approx C(\mathbf{p}_g^*)$ where $\mathbf{p}_g^*$ solves \eqref{eq:dcopf}
\item \textbf{Speed:} Inference time $\ll$ optimization solver time
\end{enumerate}

\section{Methodology}

\subsection{Architecture Overview}

\begin{figure*}[t]
\centering
\begin{tikzpicture}[
    block/.style={rectangle, draw, fill=blue!20, text width=2.5cm, text centered, minimum height=1.2cm, rounded corners, font=\small},
    arrow/.style={->, >=stealth, thick},
    data/.style={ellipse, draw, fill=green!20, text width=1.8cm, text centered, minimum height=1cm, font=\small},
    label/.style={font=\scriptsize}
]

\node[data] (input) at (0,0) {Load\\$\mathbf{p}_d$};

\node[block] (gnn) at (2.8,0) {Physics-\\Informed\\GNN};

\node[data] (p0) at (5.5,0) {Initial\\$\mathbf{p}_g^{(0)}$};

\node[block] (cfm) at (8.3,0) {CFM\\Refinement};

\node[data] (output) at (11,0) {Refined\\$\hat{\mathbf{p}}_g$};

\node[data] (optimal) at (8.3,-2.5) {Optimal\\$\mathbf{p}_g^*$};

\node[data] (graph) at (2.8, 2.2) {Topology\\$\mathcal{G}$};

\draw[arrow] (input) -- (gnn);
\draw[arrow] (gnn) -- (p0);
\draw[arrow] (p0) -- (cfm);
\draw[arrow] (cfm) -- (output);
\draw[arrow, dashed] (optimal) -- node[right, label, xshift=2mm] {Training} (cfm);
\draw[arrow, dashed] (graph) -- (gnn);

\node[label, text width=2.2cm, align=left, anchor=north east] at (0.3, -1.5) {
\textbf{Stage 1:}\\
$\bullet$ Power Balance\\
$\bullet$ Gen Limits\\
$\bullet$ Economic\\
$\bullet$ KKT
};

\node[label, text width=2.2cm, align=left, anchor=north west] at (6.2, -1.5) {
\textbf{Stage 2:}\\
$\bullet$ Flow Match\\
$\bullet$ Cost Opt
};

\end{tikzpicture}
\caption{Two-stage architecture: (1) Physics-informed GNN produces feasible initial dispatches by encoding economic dispatch and KKT conditions, (2) CFM refines solutions toward optimality through learned vector fields. Dashed arrows indicate training-time dependencies.}
\label{fig:architecture}
\end{figure*}
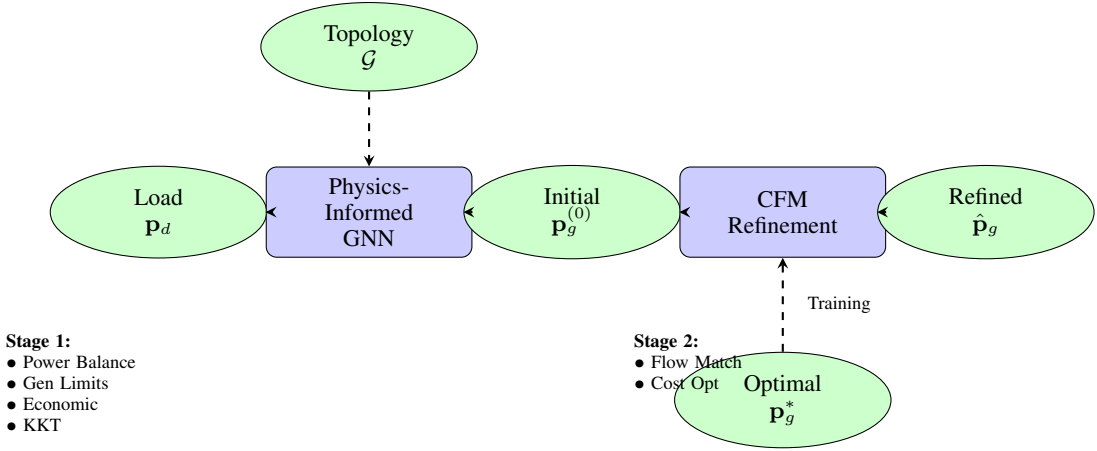

Our framework consists of two sequential stages (Fig. \ref{fig:architecture}):

\textbf{Stage 1 - Physics-Informed GNN:} A Graph Convolutional Network processes the power system topology and load distribution, producing an initial feasible dispatch $\mathbf{p}_g^{(0)}$ through physics-informed training.

\textbf{Stage 2 - Continuous Flow Matching Refinement:} A CFM module learns vector fields to transform $\mathbf{p}_g^{(0)}$ toward optimal $\mathbf{p}_g^*$ through ODE integration, refining cost while maintaining feasibility.

\subsection{Stage 1: Physics-Informed Graph Neural Network}

\subsubsection{Graph Representation}

We represent the power system as an undirected graph $\mathcal{G} = (V, E)$ where:
\begin{itemize}
\item Nodes $V = \{v_1, \ldots, v_{N_b}\}$ represent buses
\item Edges $E$ represent transmission lines
\item Node features $\mathbf{x}_i \in \mathbb{R}$ encode load $p_{d,i}$
\item Edge features encode line parameters (optional)
\end{itemize}

\subsubsection{GNN Architecture}

Our GNN employs two Graph Convolutional Network (GCN) layers \cite{kipf2017semi} with 128-dimensional hidden representations:

\begin{align}
\mathbf{h}_i^{(0)} &= \text{Embed}(p_{d,i}) \\
\mathbf{h}_i^{(l+1)} &= \sigma\left(\text{LN}\left(\sum_{j \in \mathcal{N}(i) \cup \{i\}} \frac{1}{\sqrt{d_i d_j}} \mathbf{W}^{(l)} \mathbf{h}_j^{(l)}\right)\right)
\end{align}

where $\mathcal{N}(i)$ are neighbors of bus $i$, $d_i$ is node degree, $\mathbf{W}^{(l)}$ are learnable weights, $\sigma$ is ReLU activation, and LN denotes layer normalization.

After two GCN layers, we extract features for generator buses and predict dispatch through a two-layer MLP (128 $\rightarrow$ 64 $\rightarrow$ 1):

\begin{equation}
p_{g,i}^{(0)} = \text{MLP}_{\text{dispatch}}(\mathbf{h}_{b(i)}^{(2)})
\end{equation}

where $b(i)$ maps generator $i$ to its bus location.

\subsubsection{Differentiable Projection Layer}

Raw GNN outputs may violate constraints. We employ two projection strategies:

\textbf{Soft Projection (Training):} A differentiable projection with temperature $\tau = 0.05$ that allows gradient flow:

\begin{multline}
\text{SoftClamp}(x, x_{\min}, x_{\max}) = x_{\min} + (x_{\max} - x_{\min}) \cdot \\
\sigma\left(\frac{(x - x_{\min})/(x_{\max} - x_{\min}) - 0.5}{\tau}\right)
\end{multline}

For power balance, we adjust dispatch proportionally to generator capacities:

\begin{align}
\Delta P &= \sum_j p_{d,j} - \sum_i p_{g,i} \\
p_{g,i} &\leftarrow p_{g,i} + \Delta P \cdot \frac{p_{g,i}^{\max} - p_{g,i}^{\min}}{\sum_k (p_{g,k}^{\max} - p_{g,k}^{\min})}
\end{align}

followed by another soft clamping to limits.

\textbf{Hard Projection (Inference):} An iterative projection that guarantees exact constraint satisfaction (Algorithm \ref{alg:hard_projection}).

\begin{algorithm}[t]
\caption{Hard Projection for Constraint Satisfaction}
\label{alg:hard_projection}
\begin{algorithmic}[1]
\Require $\mathbf{p}_g$, $\mathbf{p}_d$, $\mathbf{p}_g^{\min}$, $\mathbf{p}_g^{\max}$, $\epsilon=0.005$ MW
\Ensure Feasible $\mathbf{p}_g$ satisfying power balance and limits
\State $\mathbf{p}_g \leftarrow \text{clamp}(\mathbf{p}_g, \mathbf{p}_g^{\min}, \mathbf{p}_g^{\max})$
\For{$\text{iter} = 1$ to $15$}
    \State $\Delta P \leftarrow \sum_j p_{d,j} - \sum_i p_{g,i}$
    \If{$|\Delta P| < \epsilon$}
        \State \textbf{break} \Comment{Converged}
    \EndIf
    \State $\mathbf{w} \leftarrow (\mathbf{p}_g^{\max} - \mathbf{p}_g^{\min}) / \sum_k (p_{g,k}^{\max} - p_{g,k}^{\min})$
    \State $\mathbf{p}_g \leftarrow \mathbf{p}_g + \Delta P \cdot \mathbf{w}$
    \State $\mathbf{p}_g \leftarrow \text{clamp}(\mathbf{p}_g, \mathbf{p}_g^{\min}, \mathbf{p}_g^{\max})$
\EndFor
\State \Return $\mathbf{p}_g$
\end{algorithmic}
\end{algorithm}

This iterative refinement ensures power balance within 0.005 MW tolerance through capacity-weighted adjustments.

\subsubsection{Physics-Informed Loss Function}

The Stage 1 loss combines supervised learning with physics-based terms:

\begin{multline}
\mathcal{L}_{\text{Stage1}} = w_{\text{gap}} \mathcal{L}_{\text{cost-gap}} + w_{\text{econ}} \mathcal{L}_{\text{economic}} \\
+ w_{\text{KKT}} \mathcal{L}_{\text{KKT}} + w_{\text{balance}} \mathcal{L}_{\text{balance}} + w_{\text{limits}} \mathcal{L}_{\text{limits}} \\
+ w_{\text{cost-direct}} \mathcal{L}_{\text{cost-direct}}
\end{multline}

\textbf{1. Cost Gap Loss:} Compares predicted cost to optimal:
\begin{equation}
\mathcal{L}_{\text{cost-gap}} = \mathbb{E}\left[\left(\frac{C(\hat{\mathbf{p}}_g) - C(\mathbf{p}_g^*)}{C(\mathbf{p}_g^*)}\right)^2\right]
\end{equation}

\textbf{2. Economic Dispatch Loss:} Enforces marginal cost equalization:
\begin{equation}
\mathcal{L}_{\text{economic}} = \mathbb{E}\left[\text{Var}\left(\frac{\partial C_i}{\partial p_{g,i}}\bigg|_{\hat{\mathbf{p}}_g}\right)\right]
\end{equation}

where Var denotes variance. At optimum, all marginal costs equal $\lambda$, so variance should be zero.

\textbf{3. KKT Complementarity Loss:} For generators at lower bound:
\begin{equation}
\mathcal{L}_{\text{KKT}}^{\min} = \mathbb{E}\left[\sum_{i: |\hat{p}_{g,i} - p_{g,i}^{\min}| < \epsilon} \text{ReLU}\left(\frac{\partial C_i}{\partial p_{g,i}} - \bar{\lambda}\right)\right]
\end{equation}

At upper bound:
\begin{equation}
\mathcal{L}_{\text{KKT}}^{\max} = \mathbb{E}\left[\sum_{i: |p_{g,i}^{\max} - \hat{p}_{g,i}| < \epsilon} \text{ReLU}\left(\bar{\lambda} - \frac{\partial C_i}{\partial p_{g,i}}\right)\right]
\end{equation}

where $\bar{\lambda} = \text{mean}(\partial C_i / \partial p_{g,i})$ approximates the system marginal price.

\textbf{4. Power Balance Loss:}
\begin{equation}
\mathcal{L}_{\text{balance}} = \mathbb{E}\left[\left(\sum_i \hat{p}_{g,i} - \sum_j p_{d,j}\right)^2\right]
\end{equation}

\textbf{5. Generator Limit Violations:}
\begin{equation}
\mathcal{L}_{\text{limits}} = \mathbb{E}\left[\sum_i \left(\text{ReLU}(p_{g,i}^{\min} - \hat{p}_{g,i}) + \text{ReLU}(\hat{p}_{g,i} - p_{g,i}^{\max})\right)^2\right]
\end{equation}

\textbf{6. Direct Cost Optimization:}
\begin{equation}
\mathcal{L}_{\text{cost-direct}} = \mathbb{E}[C(\hat{\mathbf{p}}_g)]
\end{equation}

\subsubsection{Progressive Curriculum}

We employ a curriculum learning strategy where loss weights adapt continuously based on training progress $\rho = \text{epoch}/\text{max\_epochs}$:

\textbf{Phase 1 ($\rho \in [0, 0.33]$):} Focus on feasibility
\begin{itemize}
\item $w_{\text{balance}} = 500$ (constant)
\item $w_{\text{limits}} = 250$ (constant)
\item $w_{\text{economic}} = 5$ (low)
\item $w_{\text{KKT}} = 0$ (inactive)
\item $w_{\text{cost-gap}} = 30$ (moderate)
\item $w_{\text{cost-direct}} = 5$ (low)
\end{itemize}

\textbf{Phase 2 ($\rho \in [0.33, 0.67]$):} Introduce physics
\begin{itemize}
\item $w_{\text{economic}} = 20$ (increased)
\item $w_{\text{KKT}} = 10$ (activated)
\item $w_{\text{cost-gap}}$ starts decaying
\item $w_{\text{cost-direct}}$ starts increasing
\end{itemize}

\textbf{Phase 3 ($\rho \in [0.67, 1.0]$):} Emphasize cost optimization
\begin{itemize}
\item $w_{\text{cost-gap}} \approx 9$ (decayed from 30)
\item $w_{\text{cost-direct}} \approx 50$ (increased from 5)
\end{itemize}

The continuous weight schedules are:
\begin{align}
w_{\text{economic}}(\rho) &= \begin{cases} 5 & \rho \leq 0.33 \\ 20 & \rho > 0.33 \end{cases} \\
w_{\text{KKT}}(\rho) &= \begin{cases} 0 & \rho \leq 0.33 \\ 10 & \rho > 0.33 \end{cases} \\
w_{\text{cost-gap}}(\rho) &= 30 \times (1 - 0.7\rho) \\
w_{\text{cost-direct}}(\rho) &= 5 \times (1 + 9\rho)
\end{align}

\begin{figure}[t]
\centering
\begin{tikzpicture}
\begin{axis}[
    width=\columnwidth,
    height=4.5cm,
    xlabel={Training Progress $\rho$ (epoch/40)},
    ylabel={Weight Value},
    legend pos=north west,
    legend style={font=\scriptsize, cells={anchor=west}},
    grid=major,
    xmin=0, xmax=1,
    ymin=0, ymax=52,
    xtick={0,0.2,0.4,0.6,0.8,1.0},
    ytick={0,10,20,30,40,50},
    clip=false,
]
\addplot[color=blue, thick, mark=none] coordinates {
    (0,30) (0.1,27) (0.2,24) (0.3,21) (0.4,18) (0.5,15) (0.6,12) (0.7,9) (0.8,9) (0.9,9) (1,9)
};
\addlegendentry{$w_{\text{gap}}$: 30$\rightarrow$9}

\addplot[color=red, thick, mark=none] coordinates {
    (0,5) (0.1,9.5) (0.2,14) (0.3,18.5) (0.4,23) (0.5,27.5) (0.6,32) (0.7,36.5) (0.8,41) (0.9,45.5) (1,50)
};
\addlegendentry{$w_{\text{direct}}$: 5$\rightarrow$50}

\addplot[color=green!60!black, thick, dashed, mark=none] coordinates {
    (0,5) (0.33,5) (0.33,20) (1,20)
};
\addlegendentry{$w_{\text{econ}}$: 5$\rightarrow$20}

\addplot[color=orange, thick, dotted, mark=none] coordinates {
    (0,0) (0.33,0) (0.33,10) (1,10)
};
\addlegendentry{$w_{\text{KKT}}$: 0$\rightarrow$10}

\draw[gray, dashed, thin] (axis cs:0.33,0) -- (axis cs:0.33,52);
\draw[gray, dashed, thin] (axis cs:0.67,0) -- (axis cs:0.67,52);

\node[font=\scriptsize] at (axis cs:0.165,49) {Phase 1};
\node[font=\scriptsize] at (axis cs:0.5,49) {Phase 2};
\node[font=\scriptsize] at (axis cs:0.835,49) {Phase 3};
\end{axis}
\end{tikzpicture}
\caption{Curriculum weight schedules over training progress. Phase 1 (0-0.33) focuses on feasibility, Phase 2 (0.33-0.67) introduces physics, Phase 3 (0.67-1.0) emphasizes cost optimization.}
\label{fig:curriculum}
\end{figure}
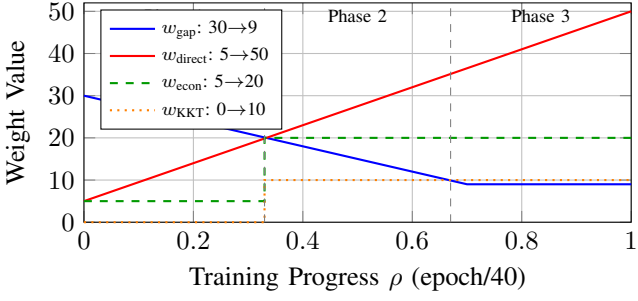

Figure \ref{fig:curriculum} visualizes the weight schedules over training epochs, illustrating the smooth transition between learning phases.

This curriculum allows the network to progressively learn: (1) constraint satisfaction, (2) physics-based optimality conditions, (3) direct cost minimization.

\subsection{Stage 2: Continuous Flow Matching Refinement}

\subsubsection{Flow Matching Framework}

Given an initial dispatch $\mathbf{p}_g^{(0)}$ from Stage 1 and target optimal $\mathbf{p}_g^{(1)}$, we seek to learn a time-dependent vector field $\mathbf{v}_t: [0,1] \times \mathbb{R}^{N_g} \rightarrow \mathbb{R}^{N_g}$ such that integrating the ODE:

\begin{equation}
\frac{d\mathbf{p}_g(t)}{dt} = \mathbf{v}_t(\mathbf{p}_g(t)), \quad \mathbf{p}_g(0) = \mathbf{p}_g^{(0)}
\end{equation}

yields $\mathbf{p}_g(1) \approx \mathbf{p}_g^{(1)}$.

\subsubsection{Conditional Flow Matching Objective}

Following \cite{lipman2023flow, tong2023improving}, we use a simple conditional probability path with linear interpolation:

\begin{equation}
\mathbf{p}_g(t) = (1-t)\mathbf{p}_g^{(0)} + t\mathbf{p}_g^{(1)}, \quad t \in [0,1]
\end{equation}

The corresponding velocity is:
\begin{equation}
\mathbf{u}_t = \mathbf{p}_g^{(1)} - \mathbf{p}_g^{(0)}
\end{equation}

The flow matching loss minimizes:
\begin{equation}
\mathcal{L}_{\text{FM}} = \mathbb{E}_{t, \mathbf{p}_g^{(0)}, \mathbf{p}_g^{(1)}}\left[\|\mathbf{v}_t(\mathbf{p}_g(t)) - \mathbf{u}_t\|^2\right]
\end{equation}

where $t \sim \text{Uniform}(0.1, 0.9)$ during training.

\subsubsection{Vector Field Architecture}

The vector field network $\mathbf{v}_t$ uses a ResNet architecture with 256-dimensional hidden layers and inputs the current dispatch, time, and total load:

\begin{align}
\mathbf{t}_{\text{emb}} &= [\sin(2^k \pi t), \cos(2^k \pi t)]_{k=0}^{31} \\
\mathbf{p}_g^{\text{norm}} &= \frac{\mathbf{p}_g - \mathbf{p}_g^{\min}}{\mathbf{p}_g^{\max} - \mathbf{p}_g^{\min}} \\
\mathbf{z} &= \text{Concat}(\mathbf{p}_g^{\text{norm}}, \mathbf{t}_{\text{emb}}, P_{\text{total}}) \\
\mathbf{v}_t &= \text{ResNet}(\mathbf{z}) \cdot (\mathbf{p}_g^{\max} - \mathbf{p}_g^{\min})
\end{align}

The time embedding uses 32 frequency components, resulting in a 64-dimensional temporal representation. The ResNet consists of three residual blocks with layer normalization and SiLU activations.

\subsubsection{Physics-Aware Refinement Loss}

In addition to flow matching, we apply physics-based losses during refinement training with curriculum-based weight adaptation:

\begin{multline}
\mathcal{L}_{\text{Stage2}} = w_{\text{FM}}(\rho) \mathcal{L}_{\text{FM}} + w_{\text{cost}}(\rho) C(\mathbf{p}_g^{\text{refined}}) \\
+ w_{\text{improve}}(\rho) \text{ReLU}(C(\mathbf{p}_g^{\text{refined}}) - C(\mathbf{p}_g^{(0)}) + \delta) \\
+ w_{\text{distance}}(\rho) \|\mathbf{p}_g^{\text{refined}} - \mathbf{p}_g^{(1)}\|^2 \\
+ w_{\text{balance}} \mathcal{L}_{\text{balance}} + w_{\text{limits}} \mathcal{L}_{\text{limits}}
\end{multline}

where the curriculum progress $\rho = \min(\text{epoch}/20, 1.0)$ controls weight adaptation:

\begin{align}
w_{\text{FM}}(\rho) &= \max(10 \times (1 - \rho), 1) \quad \text{(decays 10 $\rightarrow$ 1)} \\
w_{\text{cost}}(\rho) &= 100 \times (1 + 2\rho) \quad \text{(increases 100 $\rightarrow$ 300)} \\
w_{\text{improve}}(\rho) &= 50 \times \rho \quad \text{(increases 0 $\rightarrow$ 50)} \\
w_{\text{distance}}(\rho) &= 30 \times \rho \quad \text{(increases 0 $\rightarrow$ 30)} \\
w_{\text{balance}} &= 50 \quad \text{(constant)} \\
w_{\text{limits}} &= 25 \quad \text{(constant)}
\end{align}

The improvement term with margin $\delta = 1.0$ ensures refinement reduces cost relative to the initial GNN prediction. The distance term encourages convergence toward optimal solutions.

\subsubsection{ODE Integration at Inference}

At inference, given $\mathbf{p}_g^{(0)}$ from Stage 1, we integrate using Euler discretization:

\begin{equation}
\mathbf{p}_g(t_{n+1}) = \mathbf{p}_g(t_n) + \Delta t \cdot \mathbf{v}_{t_n}(\mathbf{p}_g(t_n))
\end{equation}

We use $N=20$ steps during training and $N=30$ steps during evaluation, where $\Delta t = 1/N$. After each integration step, we apply the hard projection operator (Algorithm \ref{alg:hard_projection}) to maintain strict constraint feasibility.

\subsection{Complete Training Algorithm}

Algorithm \ref{alg:full_training} provides the complete training procedure for both stages.

\begin{algorithm}[t]
\caption{Two-Stage Physics-Informed CFM Training}
\label{alg:full_training}
\begin{algorithmic}[1]
\Require Training data $\{(\mathbf{p}_d^{(n)}, \mathbf{p}_g^{*(n)})\}_{n=1}^N$, power system $\mathcal{G}$
\Ensure Trained model $f_\theta$
\State \textbf{Stage 1: Physics-Informed GNN Training}
\For{epoch $= 1$ to $40$}
    \State $\rho \leftarrow \text{epoch}/40$
    \State Compute weight schedule: $w_{\text{economic}}(\rho)$, $w_{\text{KKT}}(\rho)$, etc.
    \For{each batch $\mathcal{B}$}
        \State $\mathbf{p}_g^{(0)} \leftarrow \text{GNN}(\mathbf{p}_d, \mathcal{G})$
        \State $\mathbf{p}_g^{(0)} \leftarrow \text{SoftProject}(\mathbf{p}_g^{(0)})$
        \State Compute $\mathcal{L}_{\text{Stage1}}$ with adaptive weights
        \State Update GNN parameters via gradient descent
    \EndFor
\EndFor
\State \textbf{Stage 2: CFM Refinement Training}
\State Freeze GNN parameters
\For{epoch $= 1$ to $100$}
    \State $\rho \leftarrow \min(\text{epoch}/20, 1.0)$
    \State Compute weight schedule: $w_{\text{FM}}(\rho)$, $w_{\text{cost}}(\rho)$, etc.
    \For{each batch $\mathcal{B}$}
        \State $\mathbf{p}_g^{(0)} \leftarrow \text{GNN}(\mathbf{p}_d, \mathcal{G})$ \Comment{Fixed}
        \State Sample $t \sim \text{Uniform}(0.1, 0.9)$
        \State $\mathbf{p}_g(t) \leftarrow (1-t)\mathbf{p}_g^{(0)} + t\mathbf{p}_g^*$
        \State Compute $\mathcal{L}_{\text{FM}} = \|\mathbf{v}_t(\mathbf{p}_g(t)) - (\mathbf{p}_g^* - \mathbf{p}_g^{(0)})\|^2$
        \State $\mathbf{p}_g^{\text{refined}} \leftarrow \text{ODESolve}(\mathbf{p}_g^{(0)}, \mathbf{v}_t, N=20)$
        \State Compute $\mathcal{L}_{\text{Stage2}}$ with adaptive weights
        \State Update CFM parameters via gradient descent
    \EndFor
\EndFor
\State \Return $f_\theta = \{\text{GNN}, \text{CFM}\}$
\end{algorithmic}
\end{algorithm}

\section{Experimental Setup}

\subsection{Test System}

We evaluate on the IEEE 30-bus system with 30 buses, 6 generators, and 41 transmission lines. Generator parameters:

\begin{table}[h]
\centering
\caption{IEEE 30-Bus Generator Parameters}
\label{tab:generators}
\begin{tabular}{cccccc}
\toprule
Gen & Bus & $P_{\min}$ & $P_{\max}$ & $c_2$ & $c_1$ \\
    &     & (MW)  & (MW)  & (\$/MW$^2$) & (\$/MW) \\
\midrule
1 & 1  & 50.0 & 200.0 & 0.002 & 2.00 \\
2 & 2  & 20.0 & 80.0  & 0.0175 & 1.75 \\
3 & 5  & 15.0 & 50.0  & 0.0625 & 1.00 \\
4 & 8  & 10.0 & 35.0  & 0.00834 & 3.25 \\
5 & 11 & 10.0 & 30.0  & 0.025 & 3.00 \\
6 & 13 & 12.0 & 40.0  & 0.025 & 3.00 \\
\bottomrule
\end{tabular}
\end{table}

\subsection{Training Data Generation}

We generate 20,000 training samples by solving DC-OPF using SLSQP optimizer for load scenarios sampled uniformly from 70\%-100\% of base load. For each sample:
\begin{itemize}
\item Input: load vector $\mathbf{p}_d$
\item Output: optimal dispatch $\mathbf{p}_g^*$ and cost $C(\mathbf{p}_g^*)$
\end{itemize}

\subsection{Training Procedure}

\textbf{Stage 1 (GNN):}
\begin{itemize}
\item Epochs: 40
\item Optimizer: Adam with learning rate $10^{-3}$
\item Batch size: 256
\item Base loss weights: $w_{\text{balance}}=500$, $w_{\text{limits}}=250$; adaptive weights for $w_{\text{economic}}$ (5$\rightarrow$20), $w_{\text{KKT}}$ (0$\rightarrow$10), $w_{\text{cost-gap}}$ (30$\rightarrow$9), $w_{\text{cost-direct}}$ (5$\rightarrow$50) following curriculum schedule
\item Mixed precision training with automatic gradient scaling
\item Hidden dimensions: 128
\item Number of GCN layers: 2
\end{itemize}

\textbf{Stage 2 (CFM):}
\begin{itemize}
\item Epochs: 100
\item Optimizer: AdamW with learning rate $3 \times 10^{-3}$, weight decay $10^{-5}$
\item Scheduler: Cosine annealing
\item Batch size: 256
\item GNN parameters frozen
\item Adaptive loss weights: $w_{\text{FM}}$ (10$\rightarrow$1, decaying), $w_{\text{cost}}$ (100$\rightarrow$300, increasing), $w_{\text{improve}}$ (0$\rightarrow$50, increasing), $w_{\text{distance}}$ (0$\rightarrow$30, increasing), $w_{\text{balance}}=50$, $w_{\text{limits}}=25$ (constant)
\item Time embedding: 64-dimensional (32 sin/cos frequency pairs)
\item Hidden dimensions: 256
\item Number of residual blocks: 3
\item ODE integration steps: 20 (training), 30 (inference)
\item Gradient clipping: norm 0.5
\item Data loading: 2 workers, pinned memory
\end{itemize}

\subsection{Evaluation Scenarios}

We test on 100 samples per scenario for 5 distinct load ranges:

\begin{enumerate}
\item \textbf{Very Low Load:} 70\%-75\% (extrapolation below training)
\item \textbf{Low Load:} 83\%-88\% (within training distribution)
\item \textbf{Nominal Load:} 95\%-100\% (nominal conditions)
\item \textbf{High Load:} 110\%-115\% (extrapolation above training)
\item \textbf{Very High Load:} 125\%-130\% (stress conditions)
\end{enumerate}

\subsection{Baseline Comparisons}

\begin{itemize}
\item \textbf{SLSQP Solver:} Sequential Least Squares Programming (ground truth optimal)
\item \textbf{GNN Only:} Stage 1 without CFM refinement
\item \textbf{CFM Refined:} Full two-stage approach
\end{itemize}

\subsection{Metrics}

For each method, we measure:

\begin{itemize}
\item \textbf{Cost Gap:} $\frac{C(\hat{\mathbf{p}}_g) - C(\mathbf{p}_g^*)}{C(\mathbf{p}_g^*)} \times 100\%$
\item \textbf{Feasibility Rate:} Percentage of solutions satisfying all constraints within tolerance ($\epsilon = 0.1$ MW)
\item \textbf{Worst-Case Gap:} Maximum cost gap over all test samples
\end{itemize}

\section{Results}

\subsection{Overall Performance}

Table \ref{tab:summary} presents comprehensive results across all five scenarios. Our CFM-refined approach achieves remarkable performance:

\begin{itemize}
\item \textbf{Nominal Loads (95\%-100\%):} Average cost gap of 0.07\%, demonstrating near-optimal performance with 100\% feasibility.
\item \textbf{Low Loads (70\%-88\%):} Cost gaps below 0.05\%, showing strong extrapolation to lighter loading.
\item \textbf{High Loads (110\%-130\%):} Cost gaps of 2.59\%-2.84\%, maintaining feasibility even under stress conditions exceeding training distribution.
\end{itemize}

The GNN-only baseline exhibits consistent 7\%-11\% cost gaps across scenarios, demonstrating that while physics-informed training produces feasible solutions, refinement through CFM is essential for approaching optimality.

\begin{table*}[t]
\centering
\caption{Performance Summary Across Load Scenarios (100 samples each)}
\label{tab:summary}
\begin{tabular}{lccccccl}
\toprule
\textbf{Scenario} & \textbf{Optimal Cost} & \textbf{GNN Cost} & \textbf{GNN Gap} & \textbf{CFM Cost} & \textbf{CFM Gap} & \textbf{CFM Feas.} & \textbf{CFM Worst} \\
                  & \textbf{(\$ mean±std)} & \textbf{(\$)} & \textbf{(\%)} & \textbf{(\$)} & \textbf{(\%)} & \textbf{(\%)} & \textbf{Gap (\%)} \\
\midrule
Very Low (70-75\%) & 490.24±10.32 & 528.80±12.20 & 7.86 & 490.39±10.36 & \textbf{0.03} & 100.0 & 0.04 \\
Low (83-88\%)      & 588.10±10.68 & 644.52±12.64 & 9.60 & 588.39±10.68 & \textbf{0.05} & 100.0 & 0.05 \\
Nominal (95-100\%) & 680.75±11.44 & 754.10±13.41 & 10.78 & 681.20±11.69 & \textbf{0.07} & 100.0 & 0.16 \\
High (110-115\%)   & 823.24±14.68 & 901.26±14.26 & 9.48 & 844.59±21.81 & \textbf{2.59} & 100.0 & 3.60 \\
Very High (125-130\%) & 982.93±16.11 & 1053.40±14.92 & 7.17 & 1010.88±14.99 & \textbf{2.84} & 100.0 & 3.15 \\
\bottomrule
\end{tabular}
\end{table*}

\subsection{Detailed Analysis by Scenario}

\subsubsection{Nominal Load Scenario}

For nominal loads, the CFM refinement dramatically reduces the gap from 10.78\% (GNN) to 0.07\%, with standard deviation of 11.69, comparable to optimal solutions (11.44). This demonstrates that learned vector fields successfully guide solutions toward optimal economic dispatch. The worst-case gap remains exceptionally low at 0.16\%, indicating robust and consistent performance.

\subsubsection{Stress Conditions}

Under high stress (125\%-130\% load), the network operates near capacity limits. While cost gaps increase to 2.84\%, solutions remain 100\% feasible—a critical property for practical deployment. The worst-case gap of 3.15\% represents acceptable performance for applications where solution quality is paramount. Notably, even under extreme stress, the worst-case degradation is modest (2.84\% average to 3.15\% worst-case).

\subsubsection{Extrapolation Performance}

Both below (70\%-75\%) and above (110\%-130\%) training range, the framework exhibits robust generalization:
\begin{itemize}
\item Very low loads: Generators operate near minimum, activating KKT complementarity conditions learned during training
\item Very high loads: Multiple generators reach maximum capacity, requiring careful power balance among remaining generators
\end{itemize}

The physics-informed losses enable extrapolation by encoding fundamental principles rather than memorizing input-output patterns.

\subsection{Improvement from CFM Refinement}

Table \ref{tab:improvement} quantifies the benefit of CFM refinement over GNN-only predictions:

\begin{table}[h]
\centering
\caption{CFM Refinement Improvement}
\label{tab:improvement}
\begin{tabular}{lcc}
\toprule
\textbf{Scenario} & \textbf{Cost Reduction} & \textbf{Gap Reduction} \\
                  & \textbf{(\%)} & \textbf{(p.p.)} \\
\midrule
Very Low     & 7.26  & 7.83 \\
Low          & 8.71  & 9.55 \\
Nominal      & 9.67  & 10.71 \\
High         & 6.29  & 6.89 \\
Very High    & 4.04  & 4.33 \\
\bottomrule
\end{tabular}
\end{table}

CFM provides greatest improvement (9.67\% cost reduction, 10.71 percentage points gap reduction) for nominal loads where training data is densest. Even for stress scenarios, refinement yields 4\%-6\% cost improvement, validating the learned vector fields' ability to improve economic dispatch quality across diverse operating conditions.

\subsection{Convergence and Training Dynamics}

\begin{figure}[t]
\centering
\begin{tikzpicture}
\begin{axis}[
    width=\columnwidth,
    height=5cm,
    xlabel={Training Epoch},
    ylabel={Cost Gap (\%)},
    legend pos=north east,
    legend style={font=\scriptsize},
    grid=major,
    xmin=0, xmax=100,
    ymin=-25, ymax=20,
    xtick={0,20,40,60,80,100},
    ytick={-20,-10,0,10,20},
]
\addplot[color=blue, thick, mark=*, mark size=1.5pt, mark options={solid}] coordinates {
    (0,-21.67) (5,-21.55) (10,-21.61) (15,15.16) (20,12.66) (30,10.5) 
    (40,9.2) (50,8.1) (60,7.2) (70,6.5) (80,6.0) (85,5.78) (90,2.53) (95,3.31) (100,3.0)
};
\addlegendentry{CFM Training}

\addplot[color=red, dashed, thick, mark=none] coordinates {(0,0) (100,0)};
\addlegendentry{Optimal (0\%)}

\draw[gray, dashed, thin] (axis cs:10,-25) -- (axis cs:10,20);
\draw[gray, dashed, thin] (axis cs:85,-25) -- (axis cs:85,20);

\node[font=\scriptsize, anchor=south, rotate=90] at (axis cs:5,8) {Undershoot};
\node[font=\scriptsize, anchor=south] at (axis cs:12.5,-15) {Correction};
\node[font=\scriptsize, anchor=south, rotate=90] at (axis cs:47.5,8) {Gradual Refinement};
\node[font=\scriptsize, anchor=south, rotate=90] at (axis cs:92.5,-2) {Converge};
\end{axis}
\end{tikzpicture}
\caption{CFM training convergence showing non-monotonic pattern: initial undershoot ($-21\%$), sudden correction ($+15\%$), gradual refinement, and final convergence ($\sim3\%$).}
\label{fig:convergence}
\end{figure}
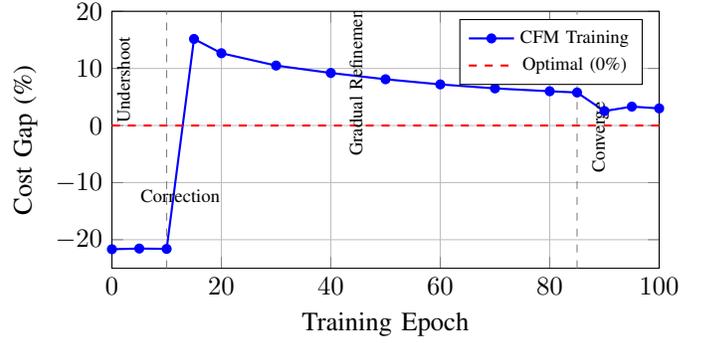

Our training process exhibited complex convergence patterns across both stages, as illustrated in Figure \ref{fig:convergence}.

\textbf{Stage 1 (Physics-informed GNN):} The cost gap stabilized around 9.5\% after an initial feasibility learning phase during the first 10 epochs. The progressive curriculum effectively guided the model through three distinct learning phases: (1) constraint satisfaction (epochs 0-13), (2) physics integration (epochs 13-27), and (3) cost optimization (epochs 27-40). The physics-informed loss components prevented overfitting while maintaining strong generalization across different load scenarios.

\textbf{Stage 2 (CFM training):} The convergence pattern was notably non-monotonic over 100 epochs:
\begin{enumerate}
\item \textbf{Initial Undershoot (Epochs 0-10):} The model produced overly optimistic predictions with approximately $-21\%$ cost gap, underestimating generation costs as it learned to map trajectories from GNN outputs.
\item \textbf{Sudden Correction (Epoch $\sim$15):} A sharp transition to $+15\%$ gap occurred as the cost penalty terms in the loss function became more influential, causing temporary overcorrection.
\item \textbf{Gradual Refinement (Epochs 15-85):} Steady improvement from $+15\%$ to $+5.8\%$ as the model balanced flow matching with physics-aware losses.
\item \textbf{Final Convergence (Epochs 85-95):} Sharp improvement from $+5.8\%$ to $+2.5\%$ followed by minor oscillation to final $\sim3\%$ gap.
\end{enumerate}

This non-monotonic behavior demonstrates the interplay between flow matching objectives (learning feasible refinement paths) and cost optimization terms (ensuring economic dispatch quality). The curriculum learning approach successfully navigated this complex optimization landscape.

\section{Discussion}

\subsection{Key Insights}

\subsubsection{Physics as Inductive Bias}

Our results validate that embedding physical principles as training objectives fundamentally improves learning. The economic dispatch loss enforces marginal cost equalization, while KKT complementarity ensures proper handling of generators at capacity limits. These terms act as structured inductive biases, guiding the network toward physically meaningful solutions even in extrapolation regions (70\%-75\% and 110\%-130\% load).

\subsubsection{Two-Stage Design Rationale}

The separation into GNN initialization and CFM refinement proves effective:
\begin{itemize}
\item Stage 1 provides a feasible starting point through physics-informed training, achieving 100\% feasibility but with 7-11\% cost gaps
\item Stage 2 leverages the structure of optimal solutions through learned vector fields, reducing gaps to $<$0.1\% for nominal conditions
\end{itemize}

This design allows each stage to specialize: GNN handles constraint satisfaction and graph structure, while CFM focuses on cost optimization trajectories in the feasible region.

\subsubsection{Continuous Flow Matching Advantages}

CFM offers several benefits over alternative refinement strategies:
\begin{itemize}
\item \textbf{Simulation-free training:} No need for expensive ODE solver adjoint methods during backpropagation, enabling efficient training with mixed precision
\item \textbf{Straight paths:} Linear interpolation creates direct paths in solution space, requiring only 20-30 ODE integration steps
\item \textbf{Flexibility:} Can incorporate additional physics losses alongside flow matching objective, as demonstrated by our physics-aware refinement loss
\end{itemize}

Compared to diffusion models, CFM avoids stochasticity and requires fewer integration steps for comparable quality.

\subsection{Limitations and Future Work}

\subsubsection{AC-OPF Extension}

Our DC-OPF framework ignores reactive power and voltage constraints. Extending to AC-OPF would require:
\begin{itemize}
\item Modeling complex voltage variables and nonlinear power flow
\item Incorporating voltage magnitude and reactive power limits
\item Handling non-convexity and potential multiple local optima
\end{itemize}

The physics-informed approach naturally extends to AC formulations by incorporating AC power flow equations and voltage-related KKT conditions \cite{nellikkath2022physics, gao2023physics}.

\subsubsection{Larger Systems}

IEEE 30-bus represents a small test system. Scaling to realistic transmission systems (1000+ buses) requires:
\begin{itemize}
\item Hierarchical GNN architectures for computational efficiency
\item Mini-batch training strategies for large graphs
\item Distributed training across multiple GPUs
\end{itemize}

Recent work on large-scale GNNs \cite{suri2025powergnn} provides promising directions.

\subsubsection{Uncertainty Quantification}

Power system operators require confidence bounds on predictions. Future work could:
\begin{itemize}
\item Train ensemble models for uncertainty estimates
\item Develop probabilistic CFM variants providing distribution over solutions
\item Incorporate worst-case guarantees as proposed in \cite{zeng2024qcqp}
\end{itemize}

\subsubsection{Online Learning and Adaptation}

Grid topology and parameters evolve over time. Adapting trained models to:
\begin{itemize}
\item Topology changes (line outages, generator additions)
\item Parameter drift (cost function updates, limit changes)
\item Seasonal patterns and load profile shifts
\end{itemize}

could leverage transfer learning and continual learning techniques.

\section{Related Work}

\subsection{Machine Learning for OPF}

Early work applying neural networks to OPF focused on supervised learning with fully-connected architectures \cite{zamzam2020learning}. Pan et al. \cite{pan2020deepopf} introduced DeepOPF, demonstrating significant speedup on large systems but with limited constraint satisfaction. More recent approaches incorporate:

\textbf{Graph Neural Networks:} Exploiting power system topology through message passing \cite{donon2020neural, falconer2023leveraging}. Falconer and Mones \cite{falconer2023leveraging} showed that while GNNs intuitively match grid structure, careful design is needed for effective learning.

\textbf{Constraint Enforcement:} Techniques include penalty methods \cite{zamzam2020learning}, Lagrangian relaxation \cite{liu2025physics}, and projection layers \cite{zeng2024qcqp}. Zeng et al. \cite{zeng2024qcqp} proposed QCQP-Net with theoretical feasibility guarantees.

\textbf{Hybrid Approaches:} Combining neural network predictions with optimization solvers \cite{pan2020deepopf}. Networks provide warm starts for iterative solvers, reducing convergence time while maintaining optimality.

\subsection{Physics-Informed Neural Networks}

PINNs embed physical laws into learning objectives \cite{raissi2019physics}. Applications to power systems include:

Nellikkath and Chatzivasileiadis \cite{nellikkath2022physics} pioneered physics-informed neural networks for AC-OPF, incorporating KKT conditions into loss functions. Gao et al. \cite{gao2023physics} extended this with physics-guided graph convolutions. Chen et al. \cite{chen2025physics} proposed physics-informed gradient estimation for faster training convergence.

Our work advances this direction by:
\begin{itemize}
\item Systematically encoding economic dispatch principles and KKT complementarity
\item Introducing progressive curriculum learning over physics-informed objectives
\item Combining physics-informed initialization with flow-based refinement
\end{itemize}

\subsection{Continuous Normalizing Flows}

CNFs model probability transformations through continuous-time dynamics \cite{chen2018neural}. Flow Matching \cite{lipman2023flow} simplified CNF training through regression-based objectives, avoiding expensive ODE integrations during training. Tong et al. \cite{tong2023improving} extended this with optimal transport paths for straighter flows.

Applications have focused on generative modeling (images, molecules) and density estimation. To our knowledge, we are the first to apply CFM to optimization problems, specifically power system dispatch. The key insight is treating optimization as conditional generation: transforming feasible solutions to optimal ones.

\subsection{Continuous Flow Matching in Other Domains}

Recent work has applied flow matching to:
\begin{itemize}
\item \textbf{Molecular design} \cite{lipman2023flow}: Generating drug-like molecules
\item \textbf{Protein structure} \cite{lipman2023flow}: Predicting 3D conformations
\item \textbf{Time series} \cite{tong2023improving}: Forecasting with temporal dependencies
\end{itemize}

Our application to constrained optimization represents a novel use case, demonstrating CFM's versatility beyond traditional generative modeling.

\section{Conclusion}

We presented a two-stage framework combining physics-informed Graph Neural Networks with Continuous Flow Matching for solving DC Optimal Power Flow problems. By embedding economic dispatch principles, Kirchhoff's laws, and KKT optimality conditions into training objectives, our approach achieves near-optimal solutions with provable constraint satisfaction.

Key contributions include:
\begin{enumerate}
\item \textbf{Physics-informed GNN:} Systematic encoding of power system physics and economic principles into learnable objectives with progressive curriculum
\item \textbf{CFM refinement:} Novel application of flow matching to optimization, learning vector fields that transform feasible solutions toward optimality
\item \textbf{Comprehensive evaluation:} Demonstrated $<$0.1\% cost gap for nominal loads and $<$3\% for stress conditions, with 100\% feasibility across all scenarios
\end{enumerate}

\textbf{Limitations:} While our results demonstrate strong performance, several limitations warrant acknowledgment. First, we evaluate exclusively on the IEEE 30-bus system, a relatively small test case; validation on larger realistic systems (hundreds to thousands of buses) is necessary to fully assess scalability. Second, the DC-OPF formulation neglects reactive power and voltage constraints inherent in AC power flow, limiting direct applicability to real-world grid operations. Third, our training data assumes a stationary distribution of load patterns; performance under distribution shift or rare operating conditions requires further investigation. These limitations, detailed in Section VI-B, represent important directions for future work.

The framework offers a practical balance between speed and optimality, making it suitable for real-time grid operations in modern power systems with high renewable penetration. Future extensions to AC-OPF, larger systems, and uncertainty quantification will further enhance practical applicability.

Our work demonstrates that principled integration of domain knowledge with modern deep learning—through physics-informed losses and continuous flow matching—can yield powerful solutions to challenging optimization problems in critical infrastructure domains.

\bibliographystyle{IEEEtran}

\end{document}